\documentclass[11pt,letterpaper]{article}
\usepackage{emnlp2017}
\usepackage{times}
\usepackage{latexsym}

\usepackage{caption}
\usepackage{url}
\usepackage{algorithm}
\usepackage{algorithmic}
\usepackage{graphicx}
\usepackage{amsmath}
\usepackage{amsfonts}

\emnlpfinalcopy

\def\emnlppaperid{654}

\title{Dependency Grammar Induction with Neural Lexicalization and Big Training Data\Thanks{This work was supported by the National Natural Science Foundation of China (61503248).}}

\author{Wenjuan Han, Yong Jiang \and Kewei Tu\\
  {\tt \{hanwj, jiangyong ,tukw\}@shanghaitech.edu.cn}\\
  School of Information Science and Technology\\ShanghaiTech University, Shanghai, China\\}

\date{}

\begin{document}

\maketitle

\begin{abstract}
We study the impact of big models (in terms of the degree of lexicalization) and big data (in terms of the training corpus size) on dependency grammar induction.
We experimented with L-DMV, a lexicalized version of Dependency Model with Valence \cite{Klein:2004:CIS:1218955.1219016} and L-NDMV, our lexicalized extension of the Neural Dependency Model with Valence \cite{jiang-han-tu:2016:EMNLP2016}. 
We find that L-DMV only benefits from very small degrees of lexicalization and moderate sizes of training corpora. L-NDMV can benefit from big training data and lexicalization of greater degrees, especially when enhanced with good model initialization, and it achieves a result that is competitive with the current state-of-the-art.
\end{abstract}

\section{Introduction}

Grammar induction is the task of learning a grammar from a set of unannotated sentences.
In the most common setting, the grammar is unlexicalized with POS tags being the tokens, and the training data is the WSJ10 corpus (the Wall Street Journal corpus with sentences no longer than 10 words) containing no more than 6,000 training sentences ~\cite{cohen2008logistic,berg2010painless,tu2012unambiguity}. 

Lexicalized grammar induction aims to incorporate lexical information into the learned grammar to increase its representational power and improve the learning accuracy. The most straightforward approach to encoding lexical information is full lexicalization \cite{pategrammar, spitkovsky2013breaking}. A major problem with full lexicalization is that the grammar becomes much larger and thus learning is more data demanding. To mitigate this problem, Headden et al. \shortcite{headden2009improving} and Blunsom and Cohn \shortcite{blunsom2010unsupervised} used partial lexicalization in which infrequent words are replaced by special symbols or their POS tags.
Another straightforward way to mitigate the data scarcity problem of lexicalization is to use training corpora larger than the standard WSJ corpus. For example, Pate and Johnson \shortcite{pategrammar} used two large corpora containing more than 700k sentences; Marecek and Straka \shortcite{marecek2013stop} utilized a very large corpus based on Wikipedia in learning an unlexicalized dependency grammar.
Finally, smoothing techniques can be used to reduce the negative impact of data scarcity. One example is Neural DMV (NDMV) \cite{jiang-han-tu:2016:EMNLP2016} which incorporates neural networks into DMV and can automatically smooth correlated grammar rule probabilities.

Inspired by this background, we conduct a systematic study regarding the impact of the degree of lexicalization and the training data size on the accuracy of grammar induction approaches.
We experimented with a lexicalized version of Dependency Model with Valence (L-DMV) \cite{Klein:2004:CIS:1218955.1219016} and our lexicalized extension of NDMV (L-NDMV). We find that L-DMV only benefits from very small degrees of lexicalization and moderate sizes of training corpora. In comparison, L-NDMV can benefit from big training data and lexicalization of greater degrees, especially when it is enhanced with good model initialization. The performance of L-NDMV is competitive with the current state-of-the-art.

\section{Methods}

\subsection{Lexicalized DMV}

We choose to lexicalize an extended version of DMV \cite{gillenwater2010sparsity}. We adopt a similar approach to that of Spitkovsky et al. \shortcite{spitkovsky2013breaking} and Blunsom and Cohn \shortcite{blunsom2010unsupervised} and represent each token as a word/POS pair. If a pair appears infrequently in the corpus, we simply ignore the word and represent it only with the POS tag. 
We control the degree of lexicalization by replacing words that appear less than a cutoff number in the WSJ10 corpus with their POS tags. With a very large cutoff number, the grammar is virtually unlexicalized; but when the cutoff number becomes smaller, the grammar becomes closer to be fully lexicalized.
Note that our method is different from previous practice that simply replaces rare words with a special ``unknown'' symbol \cite{headden2009improving}. Using POS tags instead of the ``unknown'' symbol to represent rare words can be helpful in the neural approach introduced below in that the learned word vectors are more informative.

\subsection{Lexicalized NDMV}
  \begin{figure}[t]
  \begin{center}
  \includegraphics[width=1\columnwidth,trim=0 0 0 0,clip]{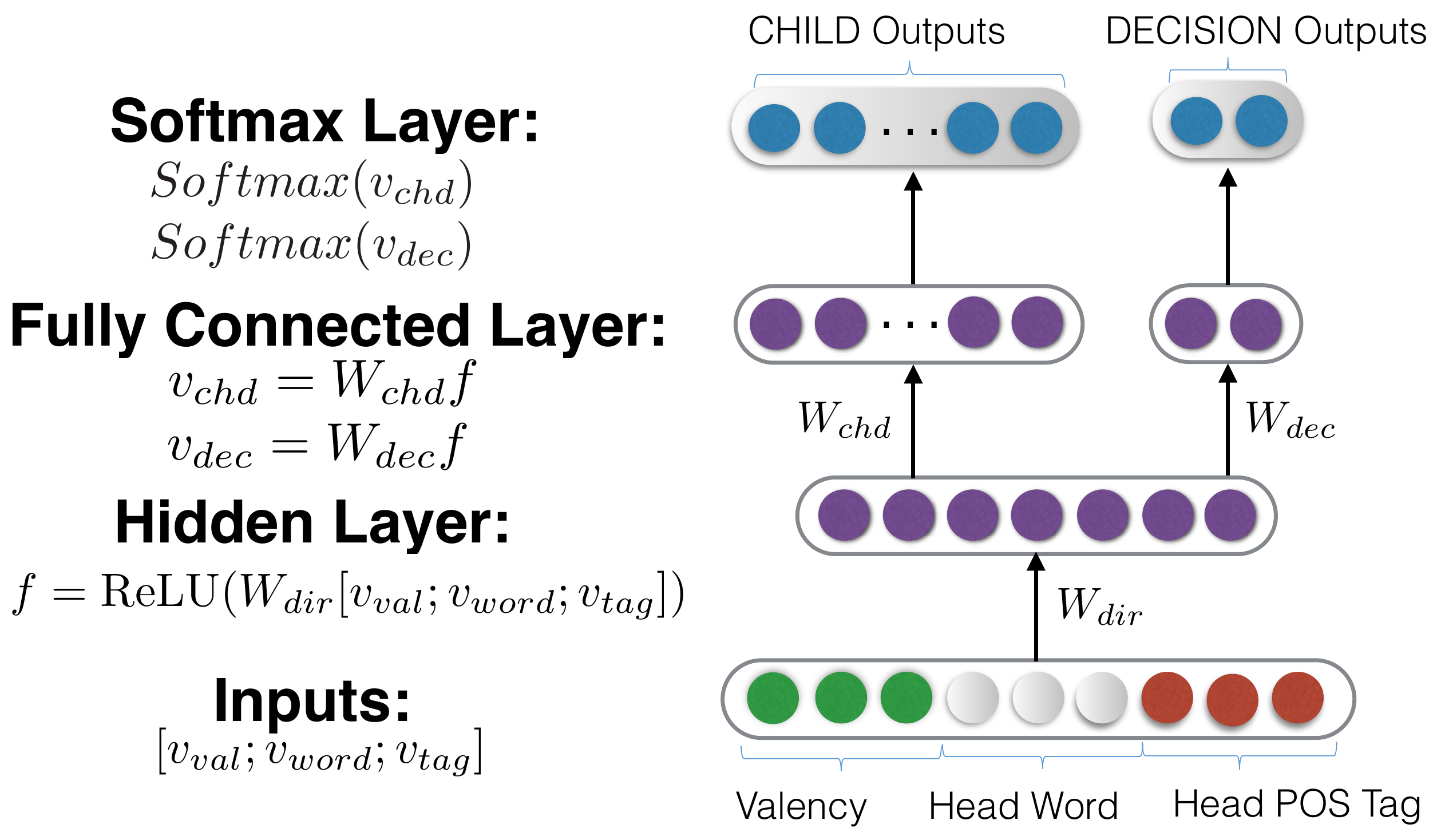}
  \caption{The structure of the neural networks in the L-NDMV model. It predicts the probabilities of the {\tt CHILD} rules and {\tt DECISION} rules.}
  \label{structure of LNDMV}
  \end{center}
  \end{figure}
With a larger degree of lexicalization, the grammar contains more tokens and hence more parameters (i.e., grammar rule probabilities), which require more data for accurate learning. Smoothing is a useful technique to reduce the demand for data in this case. Here we employ a neural approach to smoothing. Specifically, we propose a lexicalized extension of neural DMV \cite{jiang-han-tu:2016:EMNLP2016} and we call the resulting approach L-NDMV.

{\bf Extended Model:} The model structure of L-NDMV is similar to that of NDMV except for the representations of the head and the child of the {\tt CHILD} and {\tt DECISION} rules. 
The network structure for predicting the probabilities of {\tt CHILD} rules $[p_{c_{1}}, p_{c_{2}},...,p_{c_{m}}]$ ($m$ is the vocabulary size; $c_{i}$ is the $i$-th token) and {\tt DECISION} rules $[p_{stop}, p_{continue}]$ given the head word, head POS tag, direction and valence is shown in Figure \ref{structure of LNDMV}. We denote the input continuous representations of the head word, head POS tag and valence by $v_{word}$, $v_{tag}$ and $v_{val}$ respectively. By concatenating these vectors we get the input representation to the neural network: $[v_{val};v_{word};v_{tag}]$. We map the input representation to the hidden layer $f$ using the direction-specific weight matrix $W_{dir}$ and the $\mathrm{ReLU}$ activation function.
We represent all the child tokens with matrix $W_{chd}=[W_{word},W_{tag}]$ which contains two parts: child word matrix $W_{word}\in \mathbb{R}^{m \times k}$ and child POS tag matrix $W_{tag}\in \mathbb{R}^{m \times k'}$, where $k$ and $k'$ are the pre-specified dimensions of output word vectors and tag vectors respectively. The $i$-th rows of $W_{word}$ and $W_{tag}$ represent the output continuous representations of the $i$-th word and its POS tag respectively. Note that for two words with the same POS tag, the corresponding POS tag representations are the same. We take the product of $f$ and the child matrix $W_{chd}$ and apply a softmax function to obtain the {\tt CHILD} rule probabilities. For {\tt DECISION} rules, we replace $W_{chd}$ with the decision weight matrix $W_{dec}$ and follow the same procedure.

{\bf Extended Learning Algorithm:}
The original NDMV learning method is based on hard-EM and is very time-consuming when applied to L-NDMV with a large training corpus. We propose two improvements to achieve significant speedup. First, at each EM iteration we collect grammar rule counts from a different batch of sentences instead of from the whole training corpus and train the neural network using only these counts. Second, we train the same neural network across EM iterations without resetting. More details can be found in the supplementary material. Our algorithm can be seen as an extension of online EM \cite{Liang:2009:OEU:1620754.1620843} to accommodate neural network training.

\subsection{Model Initialization}\label{sec:init}

It was previously shown that the heuristic KM initialization method by Klein and Manning \shortcite{Klein:2004:CIS:1218955.1219016} does not work well for lexicalized grammar induction \cite{headden2009improving,pategrammar} and it is very helpful to initialize learning with a model learned by a different grammar induction method \cite{le2015unsupervised,jiang-han-tu:2016:EMNLP2016}.
We tested both KM initialization and the following initialization method: we first learn an unlexicalized DMV using the grammar induction method of Naseem et al. \shortcite{naseem2010using} and use it to parse the training corpus; then, from the parse trees we run maximum likelihood estimation to produce the initial lexicalized model. 

\section{Experimental Setup}\label{sec:setup}

For English, we used the BLLIP corpus\footnote{Brown Laboratory for Linguistic Information Processing (BLLIP) 1987-89 WSJ Corpus Release 1} in addition to the regular WSJ corpus in our experiments. Note that the BLLIP corpus is collected from the same news article source as the WSJ corpus, so it is in-domain and is ideal for training grammars to be evaluated on the WSJ test set. In order to solve the compatibility issue as well as improve the POS tagging accuracy, we used the Stanford tagger \cite{toutanova2003feature} to re-tag the BLLIP corpus and selected the sentences for which the new tags are consistent with the original tags, which resulted in 182244 sentences with length less than or equal to 10 after removing punctuations. We used this subset of BLLIP and section 2-21 of WSJ10 for training, section 22 of WSJ for validation and section 23 of WSJ for testing. We used training sets of four different sizes: WSJ10 only (5779 sentences) and 20k, 50k, and all sentences from the BLLIP subset. For Chinese, we obtained 4762 sentences for training from Chinese Treebank 6.0 (CTB) after converting data to dependency structures via Penn2Malt \cite{nivre2006inductive} and then stripping off punctuations. We used the recommended validation and test data split described in the documentation.

We trained the models with different degrees of lexicalization. We control the degree of lexicalization by replacing words that appear less than a cutoff number in the WSJ10 or CTB corpus with their POS tags. For each degree of lexicalization, we tuned the dimension of the hidden layer of the neural network on the validation dataset. For English, we tested nine word cutoff numbers: 100000, 500, 200, 100, 80, 70, 60, 50, and 40, which resulted in vocabulary sizes of 35, 63, 98, 166, 203, 226, 267, 306, and 390 respectively; for Chinese, the word cutoff numbers are 100000, 100, 70, 50, 40, 30, 20, 12, and 10.
Ideally, with higher degrees of lexicalization, the hidden layer dimension should be larger in order to accommodate the increased number of tokens.
For the neural network of L-NDMV, we initialized the word and tag vectors in the neural network by learning a CBOW model using the Gensim package \cite{rehurek_lrec}. We set the dimension of input and output word vectors to 100 and the dimension of input and output tag vectors to 20.
We trained the neural network with learning rate 0.03, mini-batch size 200 and momentum 0.9. 
Because some of the neural network weights are randomly initialized, the model converges to a different local minimum in each run of the learning algorithm. Therefore, for each setup we ran our learning algorithm for three times and reported the average accuracy. More detail of the experimental setup can be found in the supplementary material.

\begin{figure*}[t]
    \begin{center}
    \begin{tabular}{ccc}
    \hspace {-0.2 cm }
      \includegraphics[width=0.45\textwidth,height=0.3\textwidth]{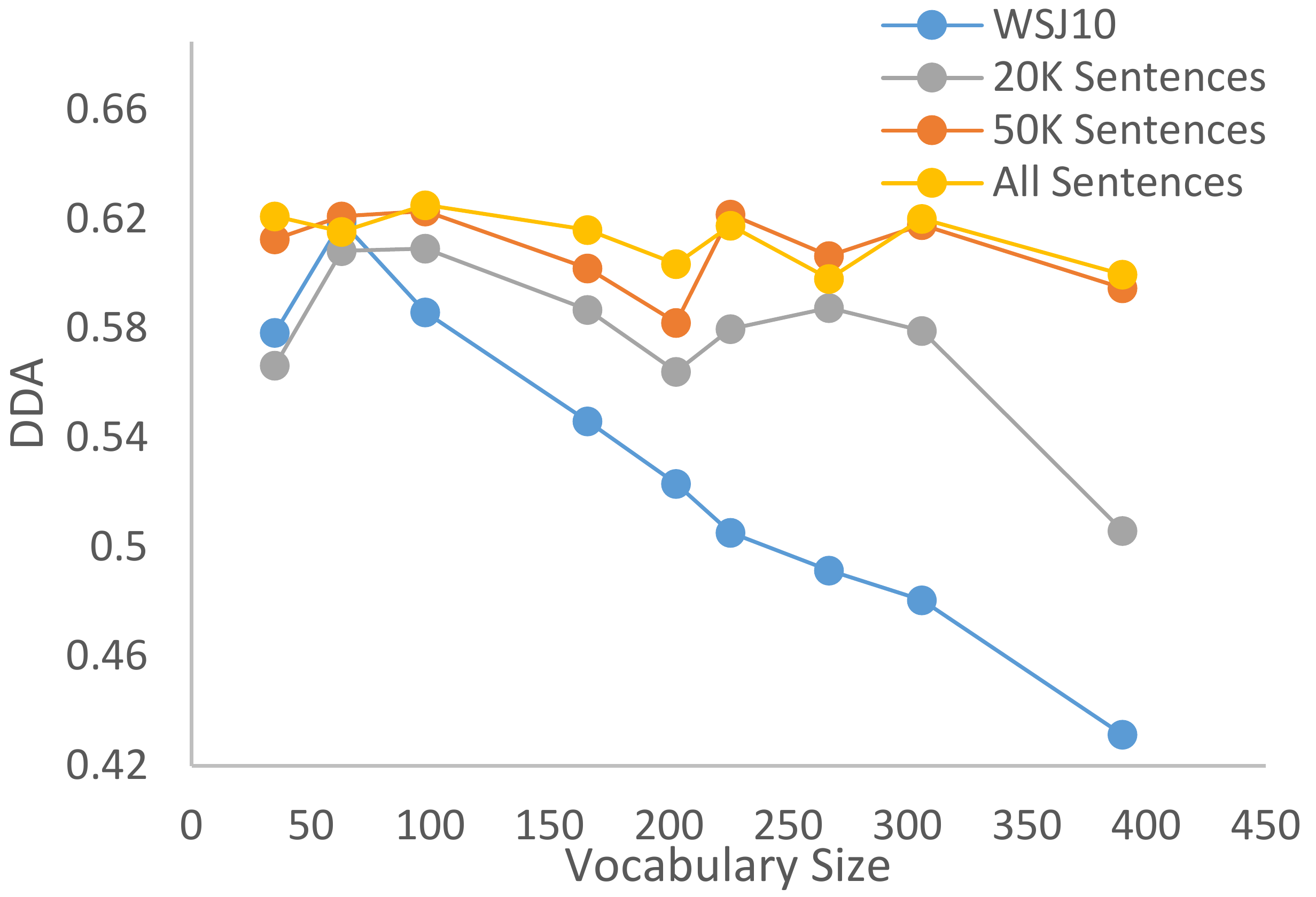} &\hspace {-0.2cm }
        \includegraphics[width=0.45\textwidth,height=0.3\textwidth]{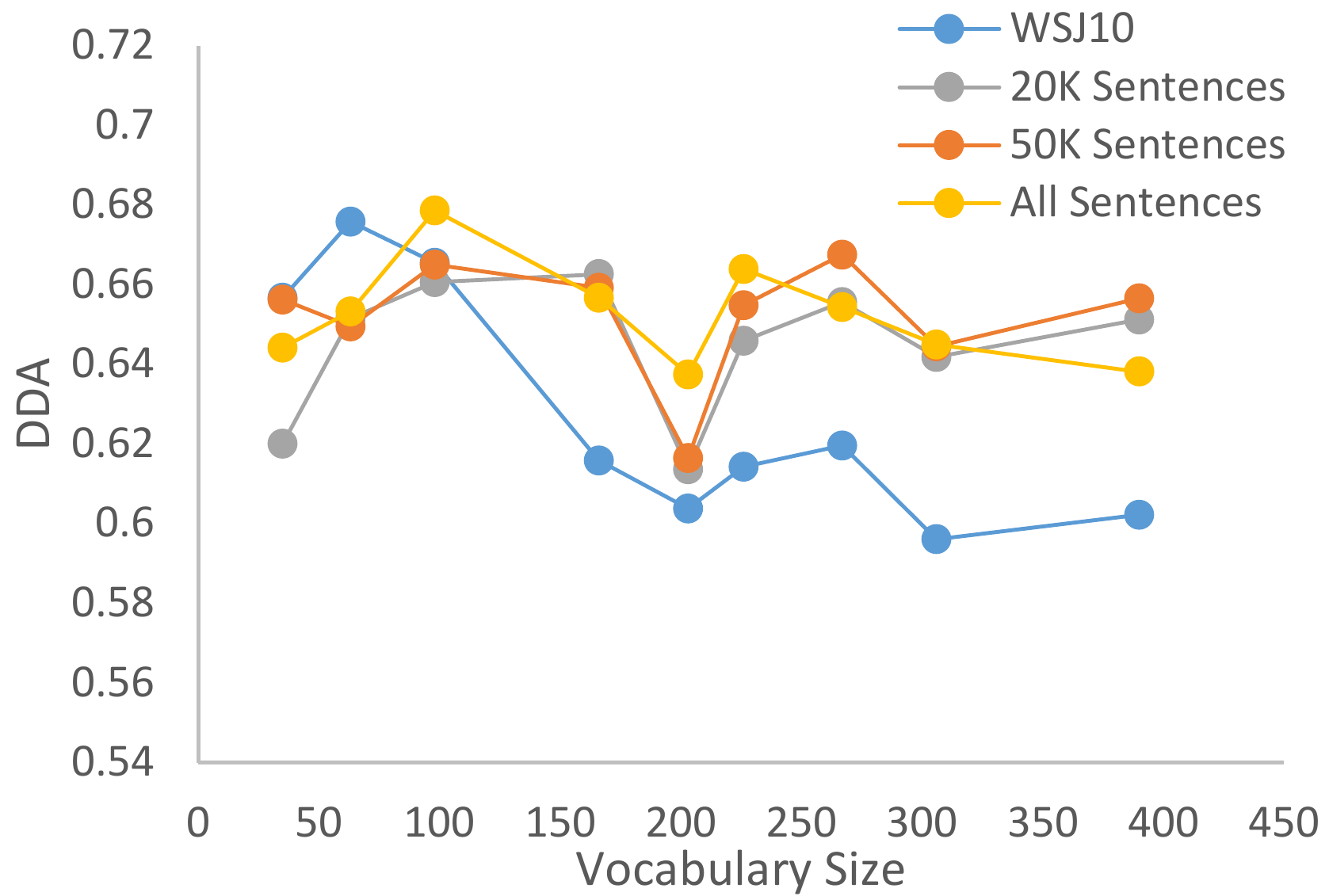} \\
     \text{\small  (a) L-DMV with KM initialization on English} & \text{\small (b) L-NDMV with KM initialization on English}\\
     \hspace {-0.2cm }
      \includegraphics[width=0.45\textwidth,height=0.3\textwidth]{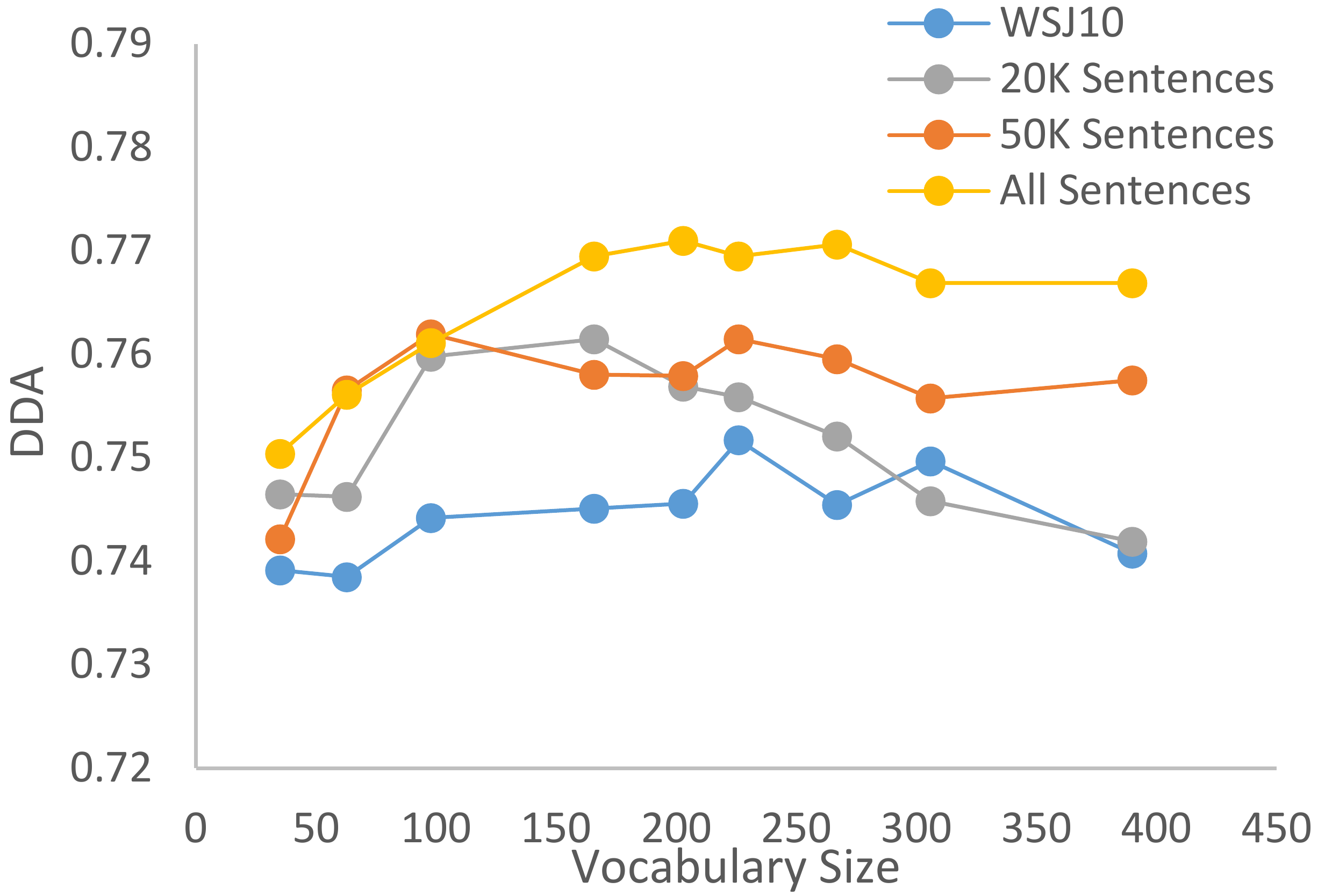}&\hspace {-0.2cm }
      \includegraphics[width=0.45\textwidth,height=0.3\textwidth]{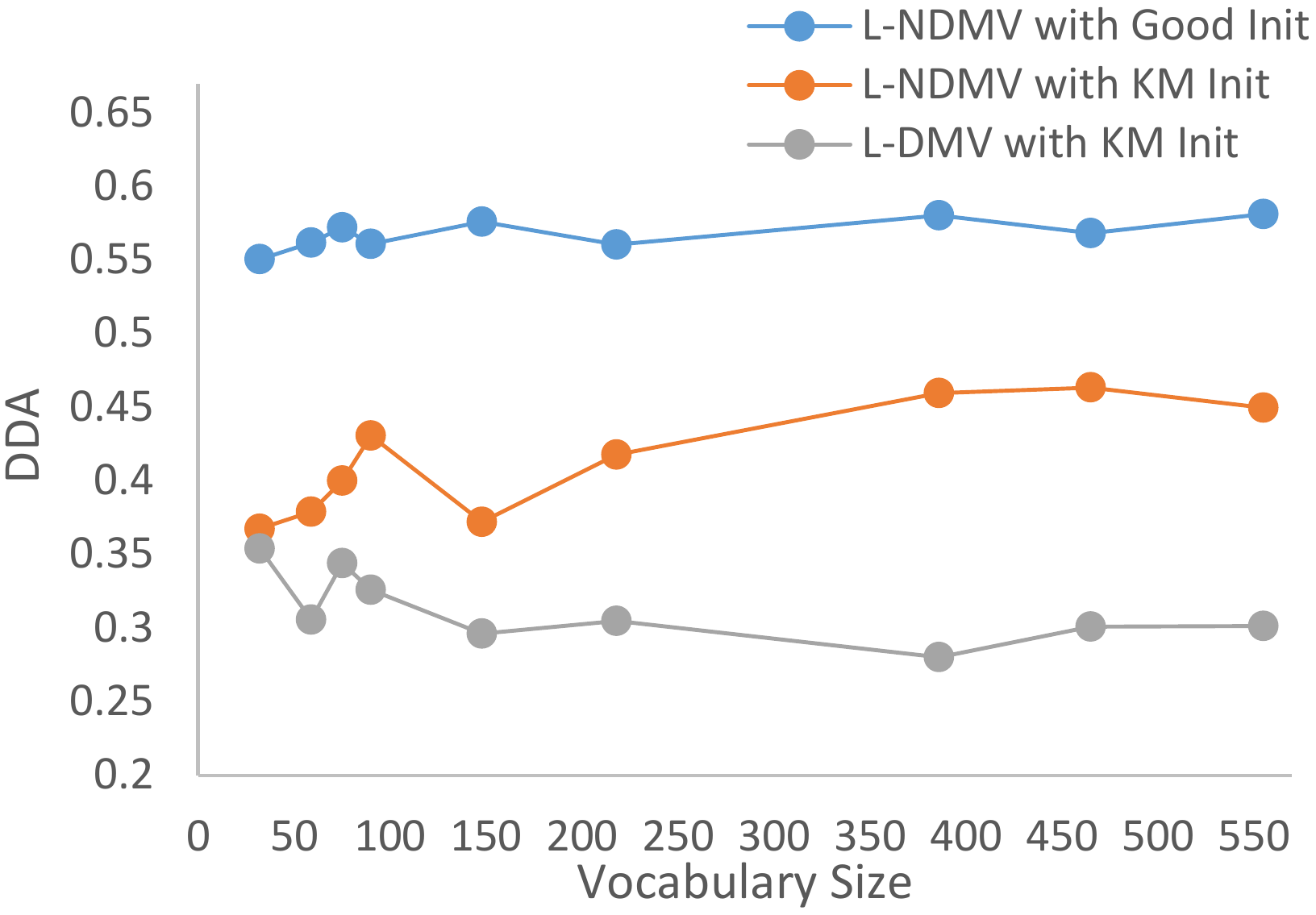}\\
     \text{\small  (c) L-NDMV with good initialization on English}  &  \text{\small (d) L-DMV and L-NDMV on Chinese}\\
    \end{tabular}
    \end{center}
    \vspace{-0.5cm}\caption{The impact of the training corpus size and the degree of lexicalization on L-DMV and L-NDMV with different initialization methods on English and Chinese.} 
    \label{fig:experimentResult}
\end{figure*}

\section{Experimental Results}
\subsection{Results on English}
Figure \ref{fig:experimentResult}(a) shows the directed dependency accuracy (DDA) of the learned lexicalized DMV with KM initialization. It can be seen that on the smallest WSJ10 training corpus, lexicalization improves learning only when the degree of lexicalization is small; with further lexicalization, the learning accuracy significantly degrades. On the three larger training corpora, the impact of lexicalization on the learning accuracy is still negative but is less severe. Overall, lexicalization seems to be very data demanding and even our largest training corpora could not bring about the benefit of lexicalization. Increasing the training corpus size is helpful regardless of the degree of lexicalization, but the learning accuracies with the 50K dataset are almost identical to those with the full dataset, suggesting diminishing return of more data.

Figure \ref{fig:experimentResult}(b) shows the results of L-NDMV with KM initialization. The parsing accuracy is improved under all the settings, showing the advantage of NDMV. The range of lexicalization degrees that improve learning becomes larger, and the degradation in accuracy with large degrees of lexicalization becomes much less severe. Diminishing return of big data as seen in the first figure can still be observed.

Figure \ref{fig:experimentResult}(c) shows the results of L-NDMV with the initialization method described in section \ref{sec:init}. It can be seen that lexicalization becomes less data demanding and the learning accuracy does not decrease until the highest degrees of lexicalization. Larger training corpora now lead to significantly better learning accuracy and support lexicalization of greater degrees than smaller corpora.
Diminishing return of big data is no longer observed, which implies further increase in accuracy with even more data.

Table \ref{table:compareRecurrentResults} compares the result of L-NDMV (with the largest corpus and the vocabulary size of 203 which was selected on the validation set) with previous approaches to dependency grammar induction. It can be seen that L-NDMV is competitive with previous state-of-the-art approaches.
We did some further analysis of the learned word vectors in L-NDMV in the supplementary material.

  \begin{table}[t]\centering
  \small
  \begin{tabular}{|l|l|l|}
  \hline
  {\bf Methods} & {\bf  WSJ10} & {\bf WSJ}\\\hline
  \hline
  \multicolumn{3}{|c|}{Unlexicalized Approaches, with WSJ10}\\ \hline%
  EVG \scriptsize \cite{headden2009improving} & 65.0 & ~~~- \\
  TSG-DMV \scriptsize \cite{blunsom2010unsupervised} & 65.9 & 53.1 \\ 
  PR-S \scriptsize \cite{gillenwater2010sparsity} & 64.3 & 53.3\\
  HDP-DEP \scriptsize \cite{naseem2010using} & 73.8 & ~~~- \\
  UR-A E-DMV \scriptsize \cite{tu2012unambiguity} & 71.4 & 57.0\\ 
  Neural E-DMV\scriptsize \cite{jiang-han-tu:2016:EMNLP2016} &  72.5 & 57.6\\\hline 
  \hline
  \multicolumn{3}{|c|}{Systems Using Lexical Information and/or More Data} \\ \hline
  LexTSG-DMV \scriptsize \cite{blunsom2010unsupervised} & 67.7 & 55.7 \\ 
  L-EVG \scriptsize \cite{headden2009improving} & 68.8 & ~~~-\\
  CS \scriptsize \cite{spitkovsky2013breaking} & 72.0 & 64.4\\
  MaxEnc \scriptsize \cite{le2015unsupervised} & 73.2 & \bf 65.8\\ \hline
  L-NDMV + WSJ \tiny & 75.1 & 59.5\\
  L-NDMV + Large Corpus & \bf 77.2 & 
  63.2 \\\hline
  \end{tabular}
  \caption{ Comparison of recent grammar induction systems.}
  \label{table:compareRecurrentResults}
  \end{table}
  
\subsection{Results on Chinese}

Figure \ref{fig:experimentResult}(d) shows the results of the three approaches on the Chinese treebank. Because the corpus is relatively small, we did not study the impact of the corpus size.
Similar to the case of English, the accuracy of lexicalized DMV degrades with more lexicalization. However, the accuracy with L-NDMV increases significantly with more lexicalization even without good model initialization. Adding good initialization further boosts the performance of L-NDMV, but the benefit of lexicalization is less significant (from 0.55 to 0.58).

\section{Effect of Grammar Rule Probability Initialization}
  \begin{figure}[t]
  \begin{center}
  \includegraphics[width=0.90\columnwidth,trim=0 0 0 0,clip]{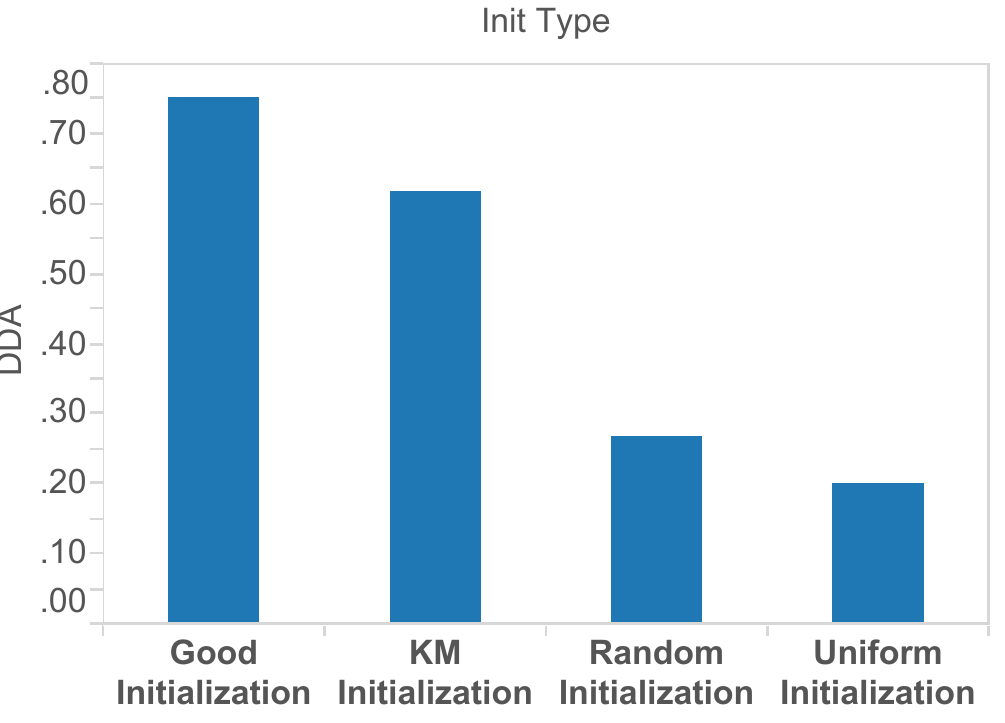}
  \caption{Comparison of four initialization methods to L-NDMV: uniform initialization, random initialization, KM initialization and good initialization.}
  \label{initTypeAffect}
  \end{center}
  \end{figure} 
    We compare four initialization methods to L-NDMV: uniform initialization, random initialization, KM initialization \cite{Klein:2004:CIS:1218955.1219016}, and good initialization as described in section 2.3 in Figure \ref{initTypeAffect}.
Here we trained the L-NDMV model on the WSJ10 corpus with the same experimental setup as in section 3.

    Again, we find that good initialization leads to better performance than KM initialization, and both good initialization and KM initialization are significantly better than random and uniform initialization.
Note that our results are different from those by Pate and Johnson \shortcite{pategrammar}, who found that uniform initialization leads to similar performance to KM initialization. We speculate that it is because of the difference in the learning approaches (we use neural networks which may be more sensitive to initialization) and the training and test corpora (we use news articles while they use telephone scripts).

\section{Conclusion and Future Work}

We study the impact of the degree of lexicalization and the training data size on the accuracy of dependency grammar induction. 
We experimented with lexicalized DMV (L-DMV) and our lexicalized extension of Neural DMV (L-NDMV). 
We find that L-DMV only benefits from very small degrees of lexicalization and moderate sizes of training corpora. In contrast, L-NDMV can benefit from big training data and lexicalization of greater degrees, especially when enhanced with good model initialization, and it achieves a result that is competitive with the state-of-the-art.

In the future, we plan to study higher degrees of lexicalization or full lexicalization, as well as even larger training corpora (such as the Wikipedia corpus). We would also like to experiment with other grammar induction approaches with lexicalization and big training data.

\bibliography{emnlp2017}
\bibliographystyle{emnlp_natbib}

\end{document}